\documentclass[letterpaper, 10 pt, conference]{ieeeconf}

\IEEEoverridecommandlockouts

\usepackage{cite}
\usepackage{amsmath,amssymb,amsfonts}
\usepackage{algorithmic}
\usepackage[pdftex]{graphicx}
\graphicspath{{./images/}, {./imgs/}}
\DeclareGraphicsExtensions{.png, .pdf}
\usepackage{textcomp}
\usepackage{xcolor}
\usepackage{dsfont}
\usepackage{microtype}
\usepackage{url}
\usepackage{eso-pic}

  \usepackage[caption=false, font=footnotesize]{subfig}

\usepackage[capitalise]{cleveref}

\DeclareMathOperator*{\argmax}{arg\,max}

\begin{document}

\title{\LARGE \bf %
Ensembles of Compact, Region-specific \& Regularized\\Spiking Neural Networks for Scalable Place Recognition
}

\author{Somayeh Hussaini\hskip5em Michael Milford\hskip5em Tobias Fischer
\thanks{The authors are with the QUT Centre for Robotics, School of Electrical Engineering and Robotics,  Queensland University of Technology, Brisbane, QLD 4000, Australia. Email: {\tt\footnotesize somayeh.hussaini@hdr.qut.edu.au}}
\thanks{This work received funding from Intel Labs to TF and MM, and from ARC Laureate Fellowship FL210100156 to MM. The authors acknowledge continued support from the Queensland University of Technology (QUT) through the Centre for Robotics.}
}

\AddToShipoutPicture*{%
    \AtTextUpperLeft{%
        \put(0,0){         %
            \begin{minipage}{\textwidth}              
                \scriptsize              
                \MakeUppercase{\newline IEEE International Conference on Robotics and Automation (ICRA) 2023. \newline Preprint version; final version will be available at} \url{https://ieeexplore.ieee.org/}  %
            \end{minipage}}     
    }
}

\maketitle

\begin{abstract}

Spiking neural networks have significant potential utility in robotics due to their high energy efficiency on specialized hardware, but proof-of-concept implementations have not yet typically achieved competitive performance or capability with conventional approaches. In this paper, we tackle one of the key practical challenges of scalability by introducing a novel modular ensemble network approach, where compact, localized spiking networks each learn and are solely responsible for recognizing places in a local region of the environment only. This modular approach creates a highly scalable system. However, it comes with a high-performance cost where a lack of global regularization at deployment time leads to hyperactive neurons that erroneously respond to places outside their learned region. Our second contribution introduces a regularization approach that detects and removes these problematic hyperactive neurons during the initial environmental learning phase. We evaluate this new scalable modular system on benchmark localization datasets Nordland and Oxford RobotCar, with comparisons to standard techniques NetVLAD, DenseVLAD, and SAD, and a previous spiking neural network system. Our system substantially outperforms the previous SNN system on its small dataset, but also maintains performance on 27 times larger benchmark datasets where the operation of the previous system is computationally infeasible, and performs competitively with the conventional localization systems.

\end{abstract}

\section{Introduction}
\label{introduction}
Spiking neural networks (SNNs) closely resemble biological neural networks~\cite{ghosh2009spiking}. Each neuron has an internal state representing its current activation, and information transfer between neurons is sparsely transmitted via spikes that occur when a neuron's internal activation exceeds a threshold~\cite{gerstner2014neuronal}. Spiking networks, when deployed on tailored neuromorphic processors, have the potential to be extremely energy efficient and process data with low latencies~\cite{frady2020neuromorphic,davies2021advancing}. 

\begin{figure*}[t]
\centering
\includegraphics[width=\linewidth]{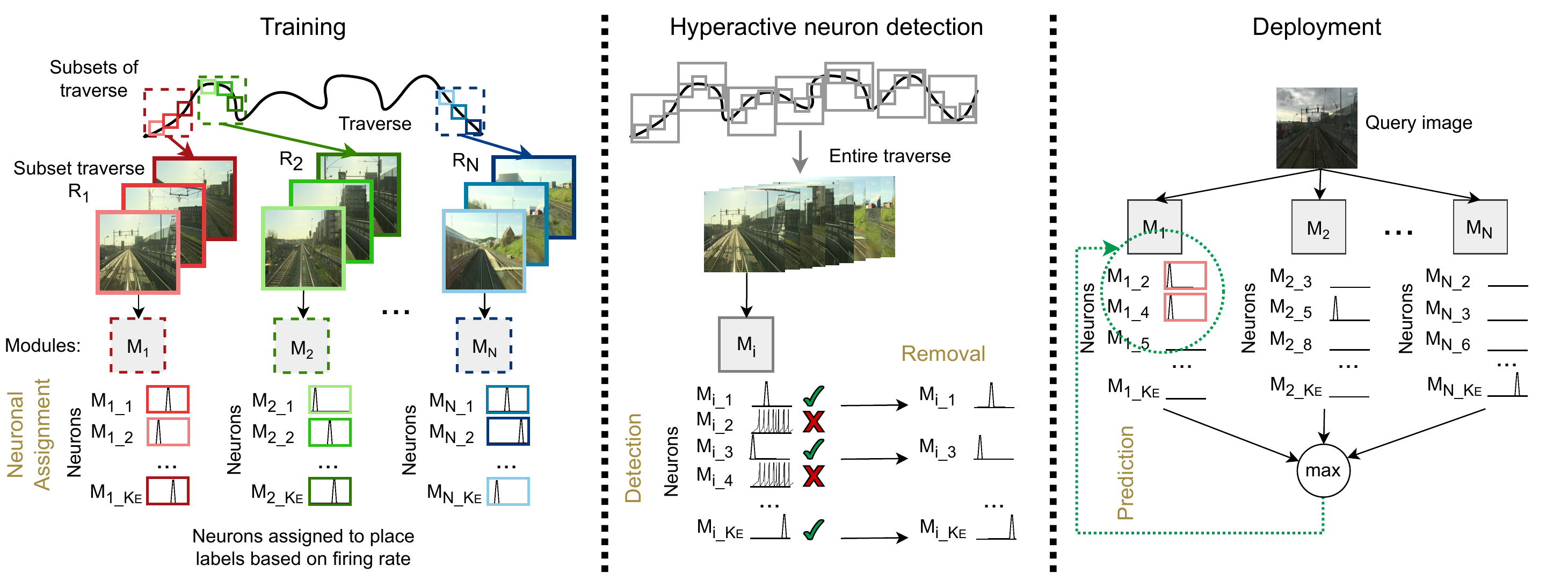}
\caption{Left: We train compact, localized spiking neural networks that solely recognize places in a local region of the environment. Middle: The independent training of these local networks leads to a lack of global regularization. This results in hyperactive neurons that strongly respond to places outside their training region. We detect and remove those hyperactive neurons. Right: At deployment time, a query image is fed to all networks in parallel. As hyperactive neurons were removed, the strongest response of \emph{all} remaining neurons in \emph{all} networks is used for the place matching decision.}
\vspace*{-0.2cm}
\label{fig:frontpage}
\end{figure*}

SNNs have thus been used in a number of robotics applications~\cite{abadia2021cerebellar, vitale2021event,dupeyroux2021neuromorphic,stagsted2020event, tieck2017towards,tieck2018controlling, kreiser2020chip, lele2021end}, including the visual place recognition (VPR) task~\cite{zhu2020spatio,Hussaini2022} that is considered in this paper. A VPR system has to find the matching reference image given a query image of a place, with the difficulty that the appearance of the query image can differ significantly from the reference image due to change in season, time of the day, or weather conditions~\cite{Garg2021,Lowry2015,masone2021survey,zhang2021visual}. VPR is crucial in a range of robot localization tasks, including loop closure detection for Simultaneous Localization and Mapping (SLAM)~\cite{Garg2021,Lowry2015,masone2021survey,zhang2021visual}.

Thus far SNNs have not been widely applied in VPR tasks. One key limitation of prior works~\cite{Hussaini2022,zhu2020spatio} is 
the specialization in \emph{only} small-scale environments. 
In this work, we aim to increase the capacity of SNNs to an order of magnitude larger environments. 
We do so by taking inspiration from the brain, which commonly uses a modular organization of neuron groups that act in parallel to efficiently perform complex recognition tasks~\cite{mountcastle1978organizing, krubitzer1995organization}. Specifically, there is evidence of an \emph{ensemble effect} for perception and learning tasks~\cite{varela2001brainweb, o2006modeling, bock2014anatomical, li2008aversive}.

In our work, we implement such an ensemble by deriving SNNs for VPR that take a divide and conquer approach \cite{jacobs1991adaptive}. Each local region of the environment is encoded in a compact, localized SNN, which is responsible only for this local region. At deployment time, all localized encoders compete with each other and are free to respond to any place, resulting in a highly scalable and parallelizable system. %
This concept is also known as a mixture of experts, where each ensemble member is an expert on a sub-task (in our case a local region of the environment), and all ensemble members cooperate to perform prediction for complex learning tasks (in our case recognizing places in a large-scale environment) \cite{jacobs1991adaptive, happel1994design}. 

We note that there are other types of ensemble learning that average the prediction of e.g.~different classifiers, but within this paper, we refer to ensembles that specialize on distinct subsets of the training data. Such \emph{independent} processing overcomes the computational constraints that arise when increasing the network size in a non-modular spiking network \cite{auda1999modular}.

While being higly scalable, localized SNNs do not interact with other ensemble members at training time and have no global regularization. As a result, some neurons erroneously respond to places outside their area of expertise. In this work, we refer to these neurons as \emph{hyperactive}. Our proposed regularization approach improves model performance by detecting and removing these problematic hyperactive neurons. %

The key contributions of our work are: 
\begin{enumerate}
    \item We introduce the concept of ensemble spiking neural networks for scalable visual place recognition (\Cref{fig:frontpage}). Each ensemble member is compact and specializes in a local region of the environment at training time. At deployment time, the query image is provided to all ensemble members in parallel, followed by a fusion of the place predictions.
    \item As each ensemble member focuses independently on a local region of the environment, there is a lack of \emph{global} regularization. After training the ensemble members, we detect hyperactive neurons, i.e.~neurons that frequently respond to images \emph{outside} their training area, and ignore the responses of these hyperactive neurons at deployment time. %
    \item We demonstrate that our method outperforms prior spiking networks~\cite{Hussaini2022} both on small datasets (for which~\cite{Hussaini2022} was designed for) and large datasets containing over 2,500 images, where~\cite{Hussaini2022} catastrophically fails. Our method performs competitively when compared to conventional VPR methods, namely NetVLAD~\cite{Arandjelovic2018}, DenseVLAD~\cite{torii201524} and SAD~\cite{milford2012seqslam}, on the Nordland~\cite{sunderhauf2013we} and Oxford RobotCar~\cite{RobotCar} datasets.
\end{enumerate}
To foster future research, we make our code available: \texttt{https://github.com/QVPR/VPRSNN}.

\section{Related works}
In this section, we review spiking neural networks in robotics research (\Cref{SNN_robotics}), ensemble neural networks concepts (\Cref{ensemble_SNN}), and key related works on visual place recognition (\Cref{bioinspired_VPR}).

\subsection{Spiking neural networks in robotics research}
\label{SNN_robotics}
The neuromorphic computing field develops hardware, sensors and algorithms that are inspired by biological neural networks, with the aim of exploiting their advantages including robustness, generalization capabilities, and incredible energy efficiency~\cite{sandamirskaya2022neuromorphic,davies2021advancing}. Spiking neural networks are one class of algorithms considered within neuromorphic computing. 

Such spiking neural networks can be trained via unsupervised methods such as Spike-Timing-Dependent-Plasticity~\cite{feldman2012spike}, or by converting pre-trained conventional artificial neural networks to spiking networks~\cite{ding2021optimal,rueckauer2017conversion,bu2021optimal}. ANN-to-SNN conversion approaches have demonstrated comparable performance to their ANN equivalents; however, these approaches typically cannot fully exploit the advantages of SNNs. %
We note that the non-differentiable nature of spikes in SNNs prevents direct application of supervised techniques such as back-propagation; however, some recent works proposed solutions to approximate back-propagation for SNNs~\cite{renner2021backpropagation, lee2020enabling}.

Thanks to their desirable characteristics, SNNs have gathered interest in a range of robotics applications, including
control~\cite{abadia2021cerebellar,vitale2021event,dupeyroux2021neuromorphic,stagsted2020event}, manipulation~\cite{tieck2017towards,tieck2018controlling}, scene understanding \cite{kreiser2020chip}, and object tracking~\cite{lele2021end}. Key works that use spiking networks for robot localization, the task considered in this paper, include 
an energy-efficient uni-dimensional SLAM~\cite{tang2019spiking},
a robot navigation controller system~\cite{tang2018gridbot},
a pose estimation and mapping system~\cite{kreiser2018pose},  
and models of the place, grid and border cells of rat hippocampus \cite{galluppi2012live} based on RatSLAM~\cite{milford2004ratslam}.

However, thus far the performance of these methods have only been demonstrated in simulated~\cite{galluppi2012live, kreiser2018pose}, constrained indoor~\cite{tang2019spiking, tang2018gridbot}, or small-scale outdoor environments~\cite{Hussaini2022}. The most similar prior work is~\cite{Hussaini2022} which introduced a high-performing SNN for VPR. However,~\cite{Hussaini2022} was limited to recognizing just 100 places, compared to several thousand places in our proposed ensemble spiking networks.

\subsection{Ensemble neural networks}
\label{ensemble_SNN}

Ensemble neural networks contain multiple ensemble members, with each ensemble member being responsible for a simple sub-task \cite{jacobs1991adaptive,happel1994design, auda1999modular}. 
In this paper, we decompose the learning data so that each ensemble member is trained in parallel on a disjoint subset of the data~\cite{auda1999modular}. %

Various ensemble schemes have been used in SNN research, including unsupervised ensembles for spiking expectation maximization networks~\cite{shim2016unsupervised}. %
The most similar ensemble SNN is that of~\cite{panda2017ensemblesnn}. However, each ensemble member in~\cite{panda2017ensemblesnn} learns a portion of an input image, opposed to different sections of the input data as in our work.

\subsection{Visual place recognition}
\label{bioinspired_VPR}
Visual place recognition (VPR) is the task of recognizing a previously visited place despite changes in appearance and perceptual aliasing \cite{Garg2021,Lowry2015}. VPR is often considered as a template matching problem, where the query image is matched to the most similar reference image.

Recent works on VPR are dominated by deep learning. A widely-known deep learning approach is NetVLAD \cite{Arandjelovic2018}, which is based on the Vector of Locally Aggregated Descriptors (VLAD) \cite{jegou2010vlad}, trained end-to-end thanks to a differentiable pooling layer. Many works extended NetVLAD in several directions~\cite{yu2019spatial, hausler2021patch, khaliq2022multires, xu2021esa}. As NetVLAD still performs competitively, we use it for benchmarking in this work. %

Bio-inspiration has a long history in VPR research: The hippocampus of rodent brains has inspired RatSLAM~\cite{milford2004ratslam}, 3D grid cells and multilayer head direction cells inspired~\cite{yu2019neuroslam}, and cognitive processes of fruit flies inspired~\cite{chancan2020hybrid}. Other works are based on spatio-temporal memory architectures~\cite{nguyen2013spatio,neubert2019neurologically}. 
We note that the detection and removal of hyperactive neurons in our approach is conceptually similar to using salient features of a place representation~\cite{newman2005slam}.

\section{Methodology}
The core idea in our method (\Cref{fig:frontpage}) is to train compact spiking networks that learn a local region of the environment (\Cref{pre}). By combining the predictions of these localized networks at deployment time within an ensemble scheme (\Cref{ensemble}) and introducing global regularization (\Cref{hyp_detection}), we enable large-scale place recognition.

\subsection{Preliminaries}
\label{pre}
Our ensemble is homogeneous, i.e.~each expert within the ensemble has the same architecture and uses the same hyperparameters. The experts only differ in their training data, which consists of geographically non-overlapping regions of the environment. The training of a single expert spiking network follows~\cite{diehl2015unsupervised,Hussaini2022} and is briefly introduced for completeness in this subsection.

\textbf{Network structure: }
Each expert module consists of three layers: 1) The input layer transforms each input image into Poisson-distributed spike trains via pixel-wise rate coding. The number of input neurons $K_P$ corresponds to the number of pixels in the input image: $K_{P} = W \times H$, where \mbox{$W$ and $H$} correspond to the width and height of the input image respectively. 
2) The $K_P$ input neurons are fully connected to $K_{E}$ excitatory neurons. Each excitatory neuron learns to represent a particular stimulus (place), and a high firing rate of an excitatory neuron indicates high similarity between the learned and presented stimuli. Note that multiple excitatory neurons can learn the same place. 3) Each excitatory neuron connects to exactly one inhibitory neuron. These inhibitory neurons inhibit all excitatory neurons except the excitatory neuron it receives a connection from. This enables lateral inhibition, resulting in a winner-takes-all system. 

\textbf{Neuronal dynamics: }
The neuronal dynamics of all neurons are implemented using the Leaky-Integrate-and-Fire (LIF) model~\cite{gerstner2014neuronal}, which describes the internal voltage of a spiking neuron in the following form: 
\begin{equation}
\tau \frac{dV}{dt} = (E_\text{rest} - V) + g_{e} (E_\text{exc} - V) + g_{i} (E_\text{inh} - V), 
\end{equation}
where $\tau$ is neuron time constant, $E_\text{rest}$ is the membrane potential at rest, $E_\text{exc}$ and $E_\text{inh}$ are the equilibrium potentials of the excitatory and inhibitory synapses with synaptic conductance $g_{e}$ and $g_{i}$ respectively.

\textbf{Network connections: }
The connections between the inhibitory and excitatory neurons are defined with constant synaptic weights. 
The synaptic conductance between input neurons and excitatory neurons is exponentially decaying, as modeled by: 
\begin{equation}
\tau_{ge} \frac{dg_{e}}{dt} = -g_{e},
\end{equation}
where the time constant of the excitatory postsynaptic neuron is $\tau_{g_{e}}$. The same model is used for inhibitory synaptic conductance $g_{i}$ with the inhibitory postsynaptic potential time constant $\tau_{g_{i}}$. 

\textbf{Weight updates: }
The biologically inspired unsupervised learning mechanism Spike-Timing-Dependent-Plasticity (STDP) is used to learn the connection weights between the input layer and excitatory neurons. Connection weights are increased if the presynaptic spike occurs before a postsynaptic spike, and decreased otherwise. %
The synaptic weight change $\Delta w$ after receiving a postsynaptic spike is defined by: 
\begin{equation}
    \Delta w = \eta (\textit{x}_\text{pre} - \textit{x}_\text{tar})(w_\text{max} - w)^\mu,
\end{equation}
where $\eta$ is the learning rate, $\textit{x}_\text{pre}$ records the number of presynaptic spikes, $\textit{x}_\text{tar}$ is the presynaptic trace target value when a postsynaptic spike arrives, $w_\text{max}$ is the maximum weight, and $\mu$ is a ratio for the dependence of the update on the previous weight.

\textbf{Local regularization: }
To prevent individual neurons from dominating the response, homeostasis is implemented through an adaptive neuronal threshold. The voltage threshold of the excitatory neurons is increased by a constant $\Theta$ after the neuron fires a spike, otherwise the voltage threshold decreases exponentially. We note that the homeostasis provides regularization only on the \emph{local}, expert-specific scale, not on the \emph{global} ensemble-level scale.

\textbf{Neuronal assignment: }
The network training encourages the network to discern the different patterns (i.e.~places) that were presented during training. As the training is unsupervised, one needs to assign each of the $K_E$ excitatory neurons to one of the $L$ training places ($K_E \gg L$). Following~\cite{diehl2015unsupervised}, we record the number of spikes $S_{e,i}$ of the $e$-th excitatory neuron when presented with an image of the $i$-th place. The highest average response of the neurons to place labels across the local training data is then used for the assignment $A_e$, such that neuron $K_{e}$ is assigned to place $l^*$ if:
\begin{equation}
    A_{e} = l^* = \argmax_l S_{e,l}
\end{equation}

\textbf{Place matching decisions: }
Following~\cite{diehl2015unsupervised}, given a query image $q$, the matched place $\hat{l}$ is the place $l$ which
is the label assigned to the group of neurons with the highest sum of spikes to the query image $\big(\text{i.e.~}A_e = \hat{l}\big)$. 
Formally:
\begin{equation}
     \hat{l} = \argmax_l \sum_{e[A_e=l]} S_{e,l}^q
     \label{eq:matching}
\end{equation}

\subsection{Ensemble Scheme} 
\label{ensemble}
The previous section described how to train individual spiking networks following~\cite{Hussaini2022}. In this section, we present our novel ensemble spiking network, which consists of a set of $\mathcal{M}=\{M_1,\dots,M_i,\dots,M_N\}$ experts. The $i$-th expert is tasked to learn the places contained in non-overlapping subsets $R_i \in \mathcal{R}$ of the reference database $\mathcal{R}$, whereby %
\begin{equation}
    \mathcal{R}=\bigcup_{i \in \{1,\dots,N\}} R_i\ \  \text{with}\ R_i \cap R_j = \varnothing\ \ \forall i\neq j.
\end{equation} 
All subsets are of equal size, i.e.~$|R_i|=\kappa$. Therefore, at training time the expert modules are independent and do not interact with each other, a key enabler of scalability.

At deployment time, the query image $q$ is provided as input to \emph{all} experts \emph{in parallel}. The place matching decision is obtained by considering the spike outputs of \emph{all} ensemble members, rather than just a single expert as in Eq.~(\ref{eq:matching}).

\subsection{Hyperactive neuron detection}
\label{hyp_detection}

The basic fusion approach that considers all spiking neurons of all ensemble members is problematic. As the expert members are only ever exposed to their local subset of the training data, there is a lack of global regularization to unseen training data outside of their local subset. In the case of spiking networks, this phenomenon leads to ``hyperactive'' neurons that are spuriously activated when stimulated with images from outside their training data. %
We decided to detect and remove these hyperactive neurons.

To detect hyperactive neurons, we do not require access to query data. We use the cumulative number of spikes $S_{e,l}^i$ fired by neurons $K_e^i$ of each module $M_i \in \mathcal{M}$ in response to the entire reference dataset $\mathcal{R}$. $S_{e,l}^i$ indicates the number of spikes fired by neuron $K_e$ of module $M_i$ in response to image $l\in\mathcal{R}$. Neuron $K_e^i$ is considered hyperactive if \begin{equation}
    \sum_l S_{e,l}^i\geq \theta,
\end{equation} 
where $\theta$ is a threshold value that is determined as described in \Cref{hyperparameter}. 
The place match is then obtained by the highest response of neurons that are assigned to place $\hat{l}$ after ignoring all hyperactive neurons:
\begin{equation}
    \hat{l} = \argmax_{i} \sum_{e[A_e=l]} S_{e,l}^q \mathds{1}_{\sum_l S_{e,l}^i<\theta},
\end{equation}
where the indicator function $\mathds{1}$ filters all hyperactive neurons.

\subsection{Hyperparameter search}
\label{hyperparameter}

We use a grid search to tune the network's hyperparameters: the time constant of the inhibitory synaptic conductance $\tau_{gi}$, and the threshold value to detect the hyperactive neurons $\theta$.  We train the modules multiple times using the reference images $\mathcal{R}$ introduced in \Cref{ensemble}, and vary $\tau_{gi}$ and $\theta$. We then observe the performance in response to a query set $\mathcal{C}$ which is geographically separate from the test set $\mathcal{T}$.

Specifically, for each combination of the hyperparameter values, we evaluate the performance of the ensemble SNN model on the $\mathcal{C}$ calibration images using the precision at 100\% recall metric (see \Cref{evaluation_metrics}). We select the $\tau_{gi}$ and $\theta$ hyperparameter values that lead to the highest performance and use these values for all test images at deployment time.

\section{Experimental Setup}

\subsection{Implementation details}
\label{imp_details}
We implemented our ensemble spiking neural network in Python3 and the Brian2 simulator~\cite{stimberg2019brian}. %
We pre-processed all reference and query input images by resizing images to $W\times H=28\times 28$ pixels, and patch-normalizing images~\cite{milford2012seqslam} using patches of size $W_P\times H_P=7\times 7$ pixels.%

We use rate coding to convert input images to Poisson spike trains. %
The number of $K_P=784$ neurons in the input layer corresponds to the number of pixels in the input image. We used $\kappa=25$ consecutive places to train each ensemble member. The hyperparameter search in \Cref{hyperparameter} resulted in $\tau_{gi}=0.5$ and $\theta=100$ for the Nordland dataset, and $\tau_{gi}=0.5$ with $\theta=180$ for the Oxford RobotCar dataset. 
Given an input image, the number of spikes of the $K_E=400$ excitatory (output) neurons in the last 10 epochs are recorded. The SNN modules were trained in parallel for 60 epochs irrespective of the dataset.

\subsection{Datasets}

We evaluated our ensemble spiking neural network on two widely used VPR datasets, Nordland \cite{sunderhauf2013we} and Oxford RobotCar \cite{RobotCar}. 
The Nordland dataset~\cite{sunderhauf2013we} captures a 728 km train journey in Norway where the same traverse is recorded during spring, summer, fall and winter. As in prior works~\cite{molloy2020intelligent, hausler2019multi, hausler2021patch} tunnels and sections where the train travels below 15 km/hr were removed. We trained our model on the spring and fall traverses, and we used summer traverse as the query dataset. We subsampled places every 8 seconds (about 100 meters) from the entire dataset, resulting in 3300 places. %
The Oxford RobotCar dataset \cite{RobotCar} contains over 100 traversals captured under varying weather conditions, times of the day and seasons. As in \cite{molloy2020intelligent}, our reference dataset consists of sun (2015-08-12-15-04-18) and rain (2015-10-29-12-18-17) traverses, and our query dataset is the dusk (2014-11-21-16-07-03) traverse. We sampled places roughly every 8 seconds (about 100 meters), resulting in 450 places.%

\subsection{Evaluation metrics}
\label{evaluation_metrics}
The precision at 100\% recall (P@100R) is the percentage of correct matches when the system is forced to match each query image to one of the reference images. The recall at $N$ (R@N) metric is the percentage of correct matches if at least one of the top $N$ predicted place labels is correctly matched. %

We consider a query image to be correctly matched only if it is matched \emph{exactly} to the correct place -- our ground truth tolerance is zero, noting that the distance between the sampled places within the datasets is relatively small.

\subsection{Baseline methods}
\label{baseline_methods}
We compare the performance of our method against three conventional VPR approaches: Firstly, the Sum-of-Absolute-Differences (SAD) \cite{milford2012seqslam} which computes the pixel-wise difference between each query image and all reference images. For a fair comparison, we applied the same resizing and patch-normalizing steps as in our approach (see \Cref{imp_details}). Secondly, DenseVLAD \cite{torii201524} which uses densely sampled SIFT image descriptors. Lastly, NetVLAD \cite{Arandjelovic2018} which generalizes across different datasets and is robust to viewpoint and appearance changes. For NetVLAD and DenseVLAD, we used the original input image size of $640 \times 360$ pixels for Nordland and resized the input images to $640 \times 480$ pixels for Oxford RobotCar, potentially giving them an advantage over the low-dimensional input images in our proposed method. 

We also compare against a non-ensemble SNN~\cite{Hussaini2022}, which in~\cite{Hussaini2022} was limited to just 100 places because of a relatively small network size. To compare against~\cite{Hussaini2022} on our large datasets, we increase the network size of their approach to contain $K_E=|\mathcal{R}|$ output neurons (i.e.~one output neuron per place). We note that increasing the number of neurons in their SNN results in significantly longer training and inference times (\Cref{fig:scalability}), so we trained their network for only 26 epochs. 
In~\Cref{small_comparison}, we additionally compare our method to~\cite{Hussaini2022} in a small-scale environment (for which~\cite{Hussaini2022} was designed).%

\begin{table}[t]
\caption{Precision at 100\% recall comparison
}
\renewcommand{\arraystretch}{1.2}
\setlength{\tabcolsep}{2.5pt}
\label{T_results}
\centering
\begin{tabular}{c|cc}

Method & Nordland & Oxford RobotCar\\\hline
Hussaini et al.~\cite{Hussaini2022} & 0.3\% & 4.0\%\\
SAD~\cite{milford2012seqslam} & 45.1\% & 41.3\%\\
DenseVLAD \cite{torii201524} & 37.9\% & \textbf{53.1}\%\\
NetVLAD~\cite{Arandjelovic2018} & 35.1\% & 44.8\%\\ 
Ensemble SNN with hyperactive neurons & 35.7\% & 30.1\%\\
Ensemble SNN (ours) & \textbf{52.6\%} & 40.5\%\\ 

\end{tabular}
\vspace{-0.2cm}
\end{table}

\section{Results}
In this section, we first provide a performance comparison of our ensemble SNN model against NetVLAD~\cite{Arandjelovic2018}, DenseVLAD~\cite{torii201524}, Sum-of-Absolute-Differences (SAD)~\cite{milford2012seqslam} and the currently best performing SNN~\cite{Hussaini2022} (\Cref{comparison}). We then evaluate the effect of removing hyperactive neurons in \Cref{detection_perf}. 
Finally, \Cref{small_comparison} provides an ablation study where we demonstrate that in small-scale environments which~\cite{Hussaini2022} was designed for, the performance of our ensemble SNN compares to prior non-ensemble SNN~\cite{Hussaini2022}.

\begin{figure}[t]
    \centering
    \includegraphics[width=0.99\linewidth,trim={1mm 1mm 1mm 1mm},clip]{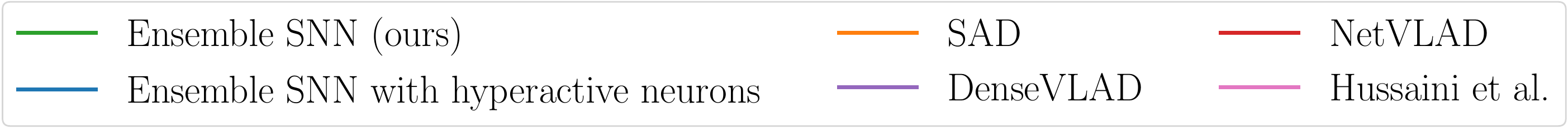}
    \subfloat[PR Nordland]{%
        \includegraphics[width=0.48\linewidth]{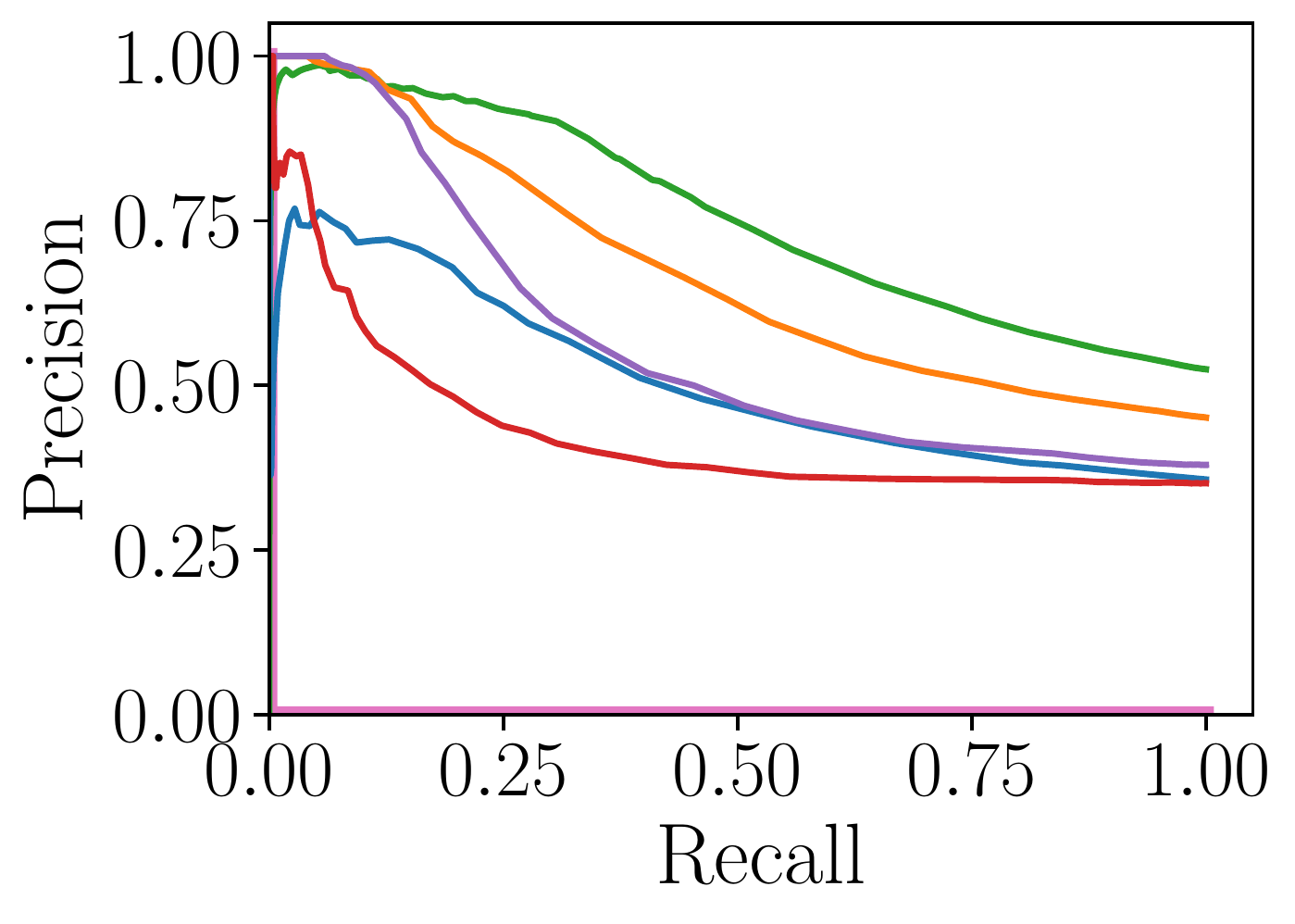}
    }
    \subfloat[PR ORC]{%
        \centering
        \includegraphics[width=0.48\linewidth]{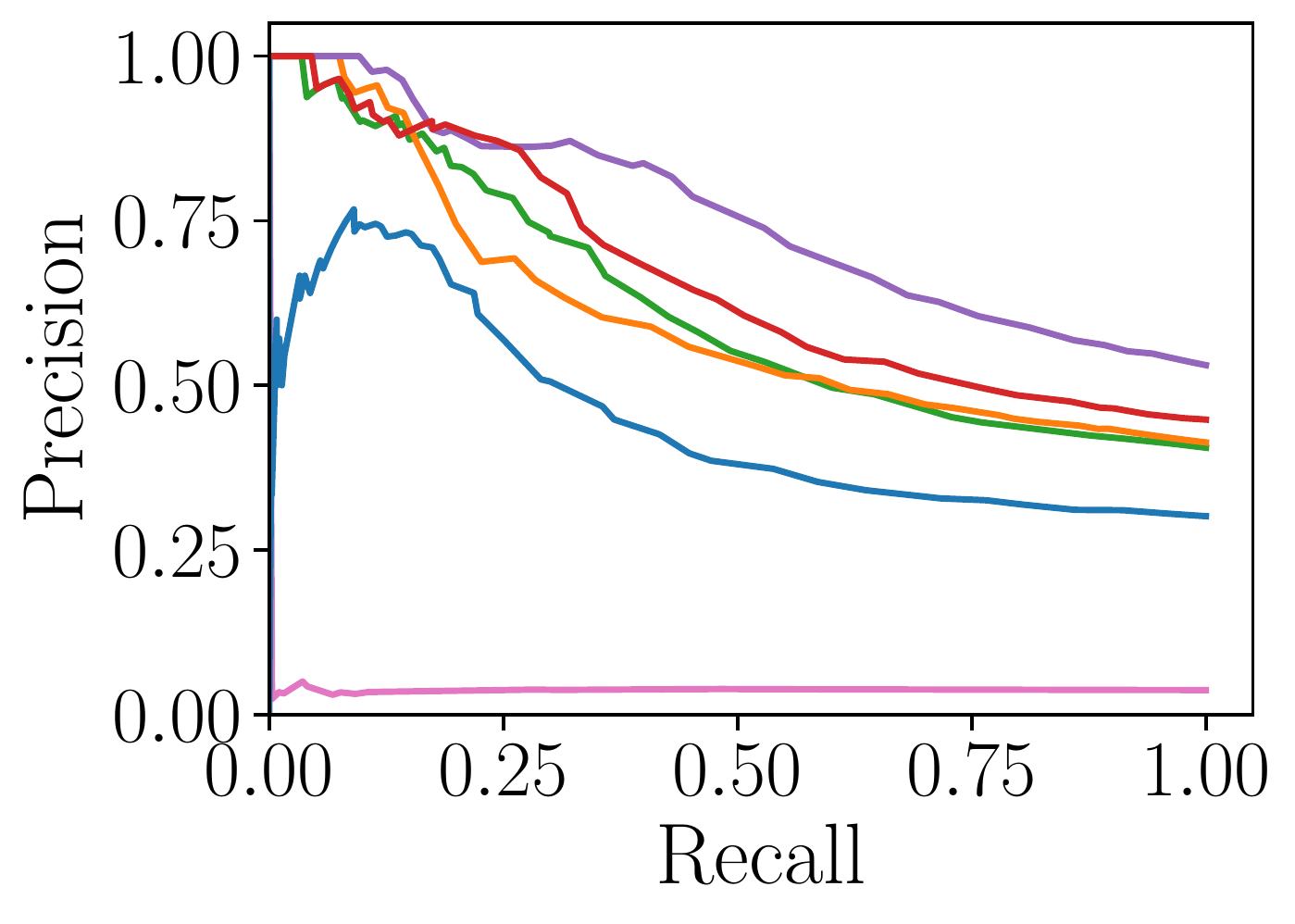}
    }\\[0.1cm]
    \hrule
    \vspace*{-0.1cm}
    \subfloat[R@N Nordland]{%
        \includegraphics[width=0.48\linewidth]{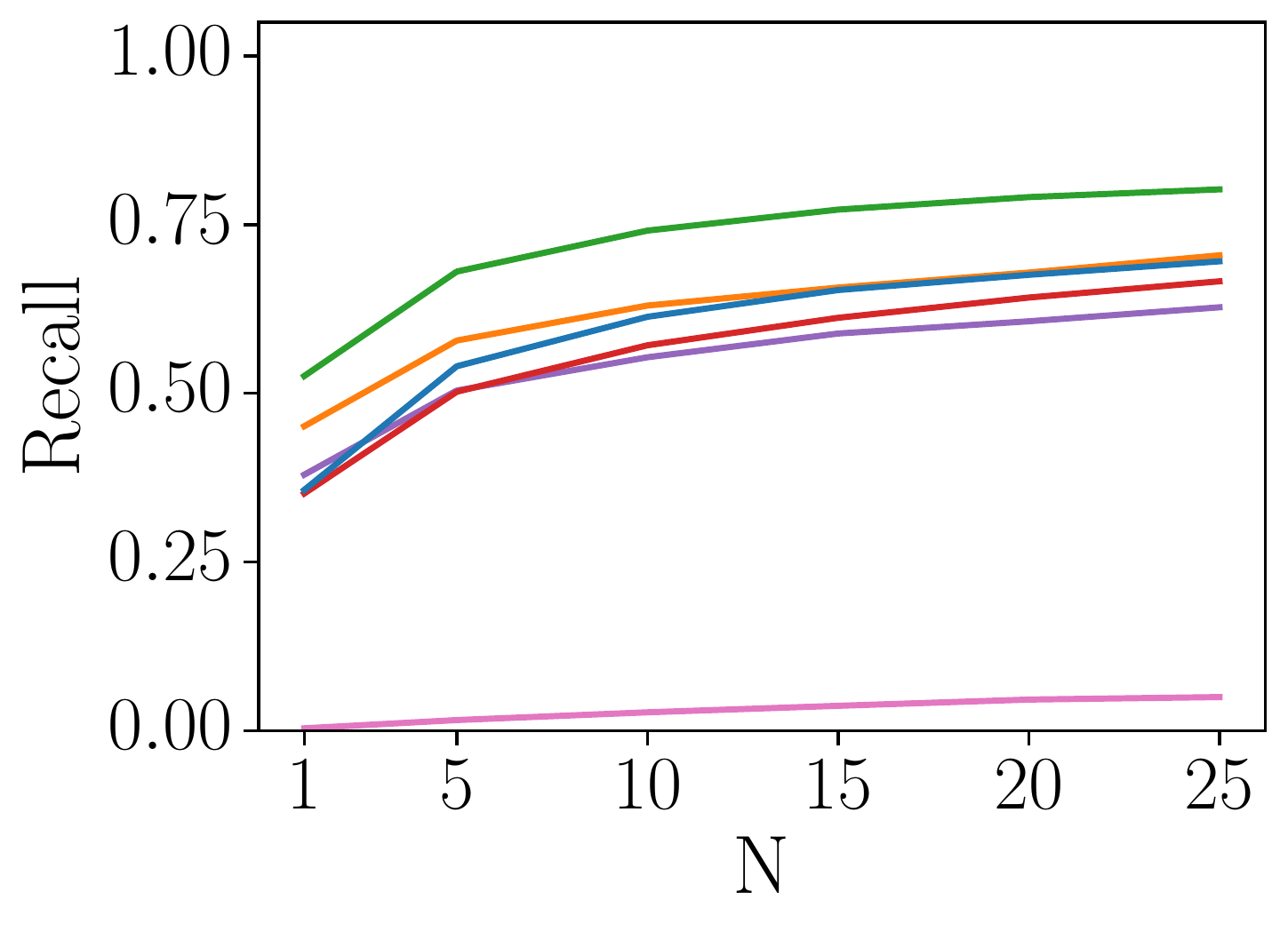}
    }
    \subfloat[R@N ORC]{%
        \includegraphics[width=0.48\linewidth]{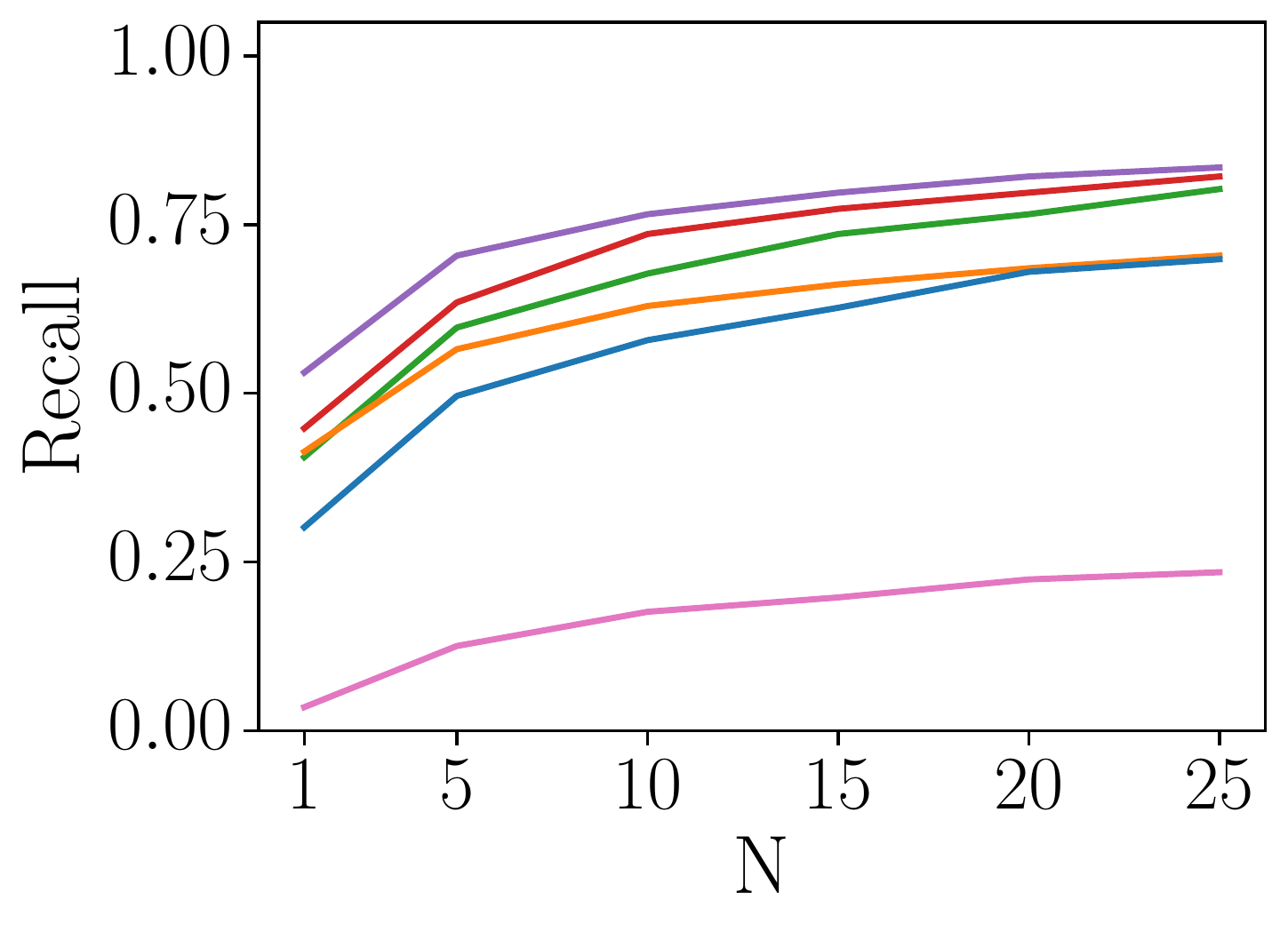}
    }
    \caption{Precision and recall curves (top) and recall@N plots (bottom) of our ensemble SNN model, ensemble SNN model where hyperactive are not ignored, the previous best performing SNN by Hussaini et al.~\cite{Hussaini2022} and conventional methods NetVLAD~\cite{Arandjelovic2018}, DenseVLAD~\cite{torii201524} and SAD~\cite{milford2012seqslam}.}
    \label{fig:PR_R@N}
    \vspace*{-0.25cm}
\end{figure}

\subsection{Comparison to state-of-the-art approaches}
\label{comparison}

We first compare our ensemble SNN to conventional VPR techniques, with the aim of merely demonstrating the potential of SNN-based approaches, as opposed to outperforming these VPR techniques. The results are summarized in \Cref{T_results}.%

For the Nordland dataset, our ensemble SNN model obtains a R@1 of 52.6\%, outperforming SAD (R@1: 45.1\%), NetVLAD (R@1: 35.1\%) and DenseVLAD (R@1: 37.9\%). %
We note that NetVLAD is known to perform relatively poorly on the rural Nordland dataset, as it was trained on urban data.
For the Oxford RobotCar dataset, the R@1 of our ensemble SNN model is 40.5\%, while the SAD approach has a similar R@1 of 41.3\%, the NetVLAD and DenseVLAD methods obtain a higher R@1 of 44.8\% and 53.1\% respectively. 
The performance of our ensemble SNN achieves a similar R@25 to NetVLAD and DenseVLAD, demonstrating the performance capability of our method.

\begin{figure}[t]
    \centering
    \includegraphics[width=0.7\columnwidth]{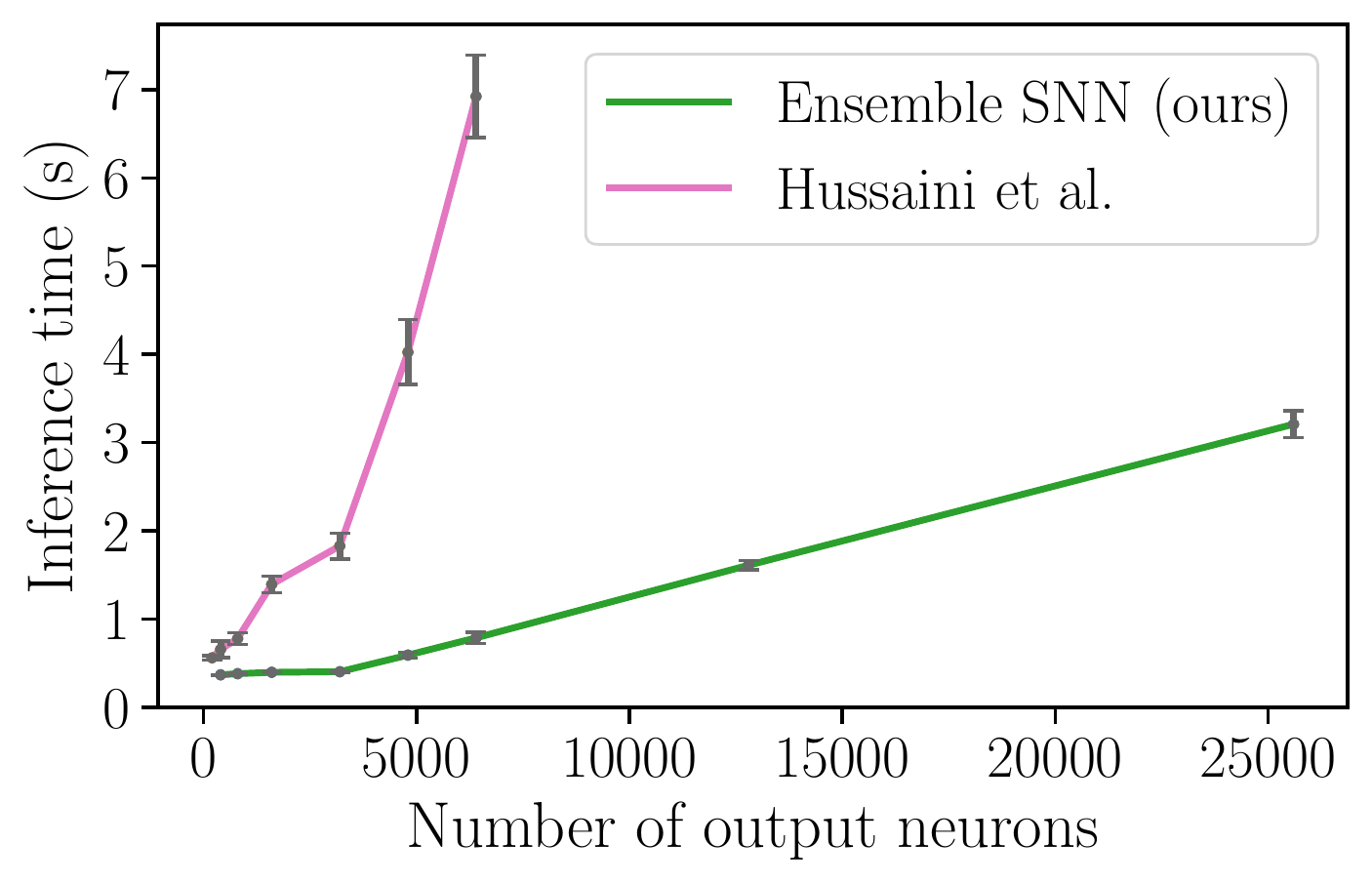}
    \vspace*{-0.25cm}
    \caption{The query time over increasing network size of our proposed ensemble SNN in comparison to~\cite{Hussaini2022}. We measured the time taken for the network to process a query image with increasing network sizes. Our ensemble SNN approach scales linearly with increasing network size. In comparison, the non-modular SNN~\cite{Hussaini2022} could only be tested for up to 6400 output neurons on CPU and its query time does not scale to large networks.}
    \label{fig:scalability}
    \vspace*{-0.1cm}
\end{figure}

\Cref{T_results} also presents the performance of the previous non-ensemble SNN model~\cite{Hussaini2022}, which is our main competitor. \cite{Hussaini2022} catastrophically failed to perform place recognition at large-scale, with a precision at 100\% recall of just 0.3\% on the Nordland and 4.0\% on the Oxford RobotCar datasets. We note that \cite{Hussaini2022} specialized in place recognition on small datasets. 
In addition to poor performance, the large number of output neurons \emph{within a single network} for~\cite{Hussaini2022} results in significantly increased inference times. This is opposed to our modular approach, where the neuronal dynamics of \emph{independent and compact} ensemble members are cheaper to compute. We highlight the computational advantages and better scalability of our method in \Cref{fig:scalability}. The inference times of our method is still slower compared to conventional VPR methods such as NetVLAD. However, deploying our method on neuromorphic hardware can significantly decrease the inference times via use of hardware parallelism. 

\subsection{Importance of hyperactivity detection}
\label{detection_perf}

\begin{figure}[t]
    \centering
        \includegraphics[width=0.49\linewidth]{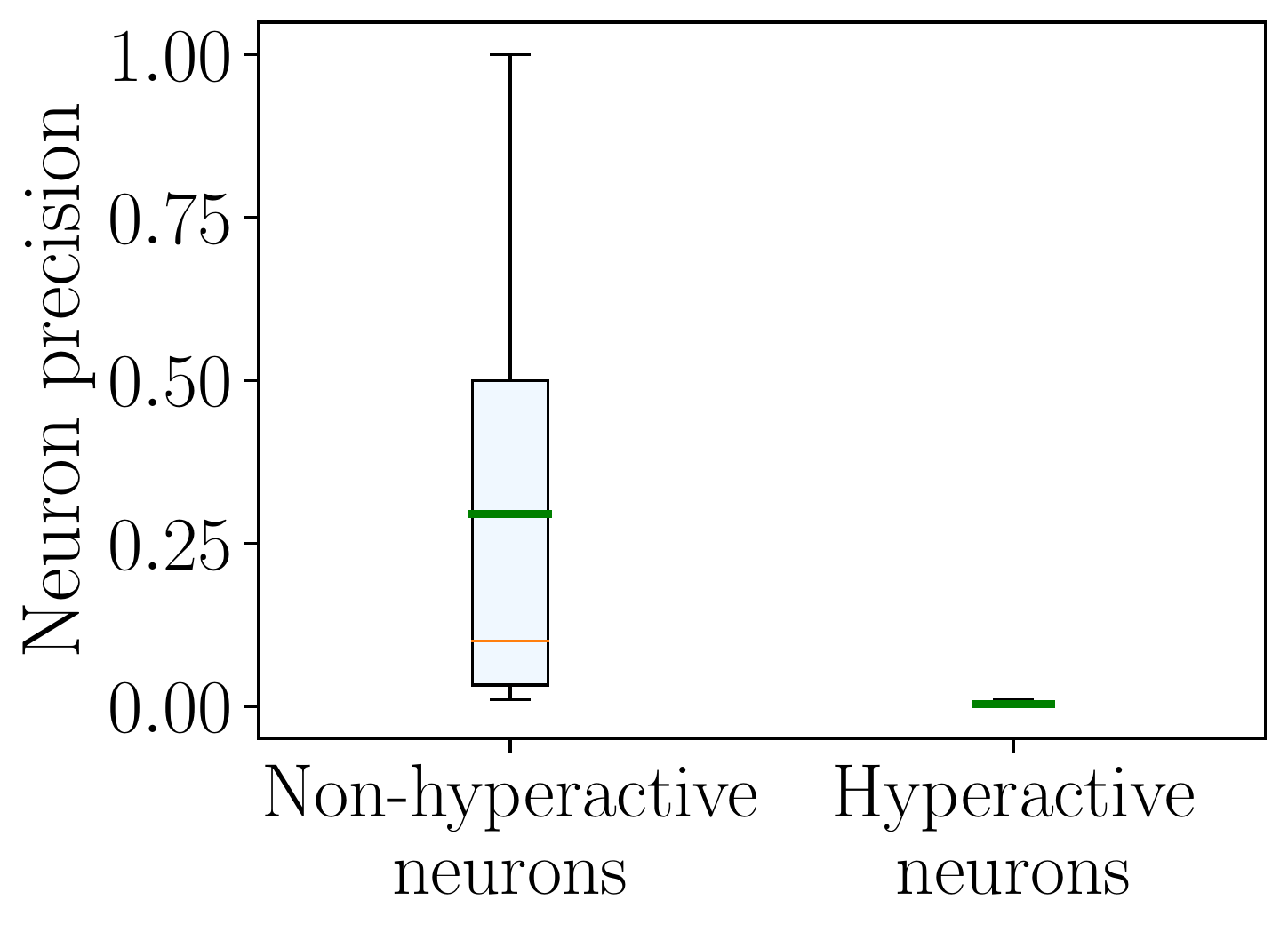}
    \vspace*{-0.25cm}
    \caption{
   We compare the neuron precision of hyperactive and non-hyperactive neurons. The neuron precision is the \textit{number of times} a neuron has fired spikes to the correct place over the \textit{total number of times} the neuron has fired spikes across the entire query dataset. The lower the neuron precision, the more (incorrect) places a neuron is responsive to, beyond the single correct place. 
    Non-hyperactive neurons have significantly higher precision in responding to correct places compared to hyperactive neurons, supporting our proposal of removing hyperactive neurons at deployment time. %
    }
    \label{fig:hyp}
    \vspace*{-0.25cm}
\end{figure}

This section evaluates that it is crucial to introduce global regularization by detecting and ignoring hyperactive neurons. 
As shown in \Cref{T_results} and \Cref{fig:PR_R@N}, our ensemble SNN model where hyperactive neurons are ignored improves the precision at 100\% recall compared to the base ensemble SNN model (that includes hyperactive neurons) on both Nordland (absolute increase of 16.9\%) and Oxford RobotCar (absolute increase of 10.4\%). %
\Cref{fig:hyp} compares the neuron precision of hyperactive and non-hyperactive neurons trained on the Nordland dataset and highlights that the precision at recognizing correct places of non-hyperactive neurons is significantly higher than that of hyperactive neurons. 

\begin{figure}[t]
    \centering
    \includegraphics[width=0.7\linewidth]{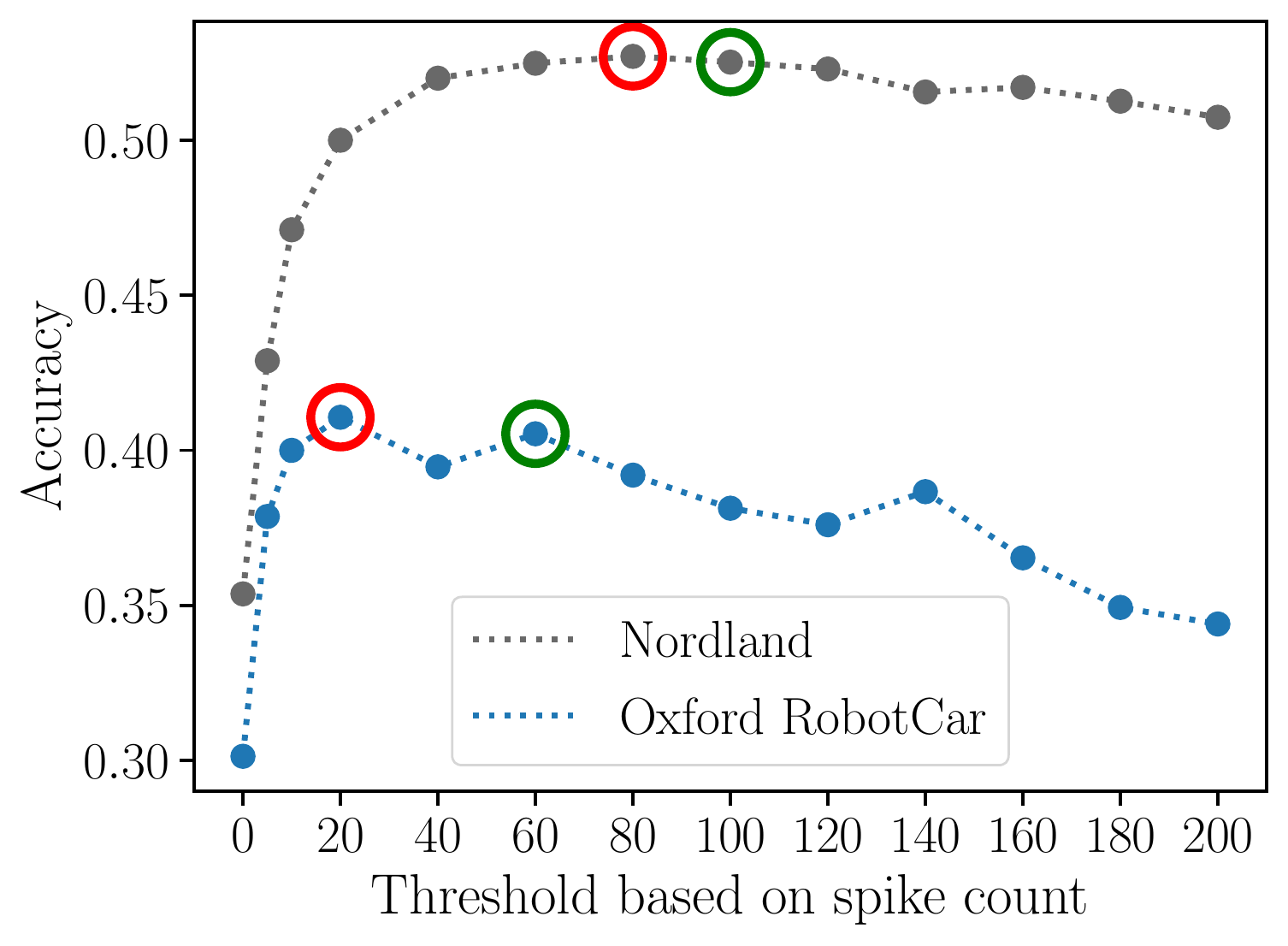}
    \vspace*{-0.2cm}
    \caption{Threshold hyperparameter selection ablation study: On the $x$-axis we plot the threshold value $\theta$ to ignore hyperactive neurons, and on the $y$-axis the precision at 100\% recall on the test set. The wide range of high-performing values indicates robustness against the precise value of $\theta$. The highest performing threshold value from the calibration process (green circle) is used to select $\theta$ for deployment. The ideal threshold that would have led to the highest performance at test time is indicated by a red circle.}
    \label{fig:abl_thr}
    \vspace*{-0.1cm}
\end{figure}

We further evaluate the sensitivity of our ensemble SNN with respect to the threshold value $\theta$ (see \Cref{hyperparameter}). 
Specifically, we evaluate the precision at 100\% recall at different threshold values ($ 0 < \theta < \theta_{max}=200$). Note that $\theta=0$ corresponds to the baseline performance where both hyperactive and non-hyperactive neurons are used. 
For both the Nordland and Oxford RobotCar datasets, \Cref{fig:abl_thr} shows that our method is not sensitive to particular values of $\theta$, with a wide range of high-performing settings. Importantly, any $\theta>0$ improves performance compared to the baseline model where hyperactive neurons are included ($\theta=0$).%

\subsection{Comparison to prior SNN in small environments}
\label{small_comparison}

This ablation presents a like-for-like comparison of our proposed ensemble SNN %
and a non-ensemble SNN model from~\cite{Hussaini2022} which facilitates direct comparison to the previous state-of-the-art VPR system using SNNs. As~\cite{Hussaini2022} was designed for small-scale datasets, we do so by considering a much smaller dataset limited to 100 places; the previous section has already shown that~\cite{Hussaini2022} fails catastrophically in large environments. %
Specifically, we trained $N=5$ ensemble members, each containing $K_E=400$ excitatory neurons. We used the first $\kappa_{cal}=25$ places to calibrate $\tau_{gi}$ and $\theta$.%

\begin{figure}[t]
    \centering
    \includegraphics[width=0.44\columnwidth]{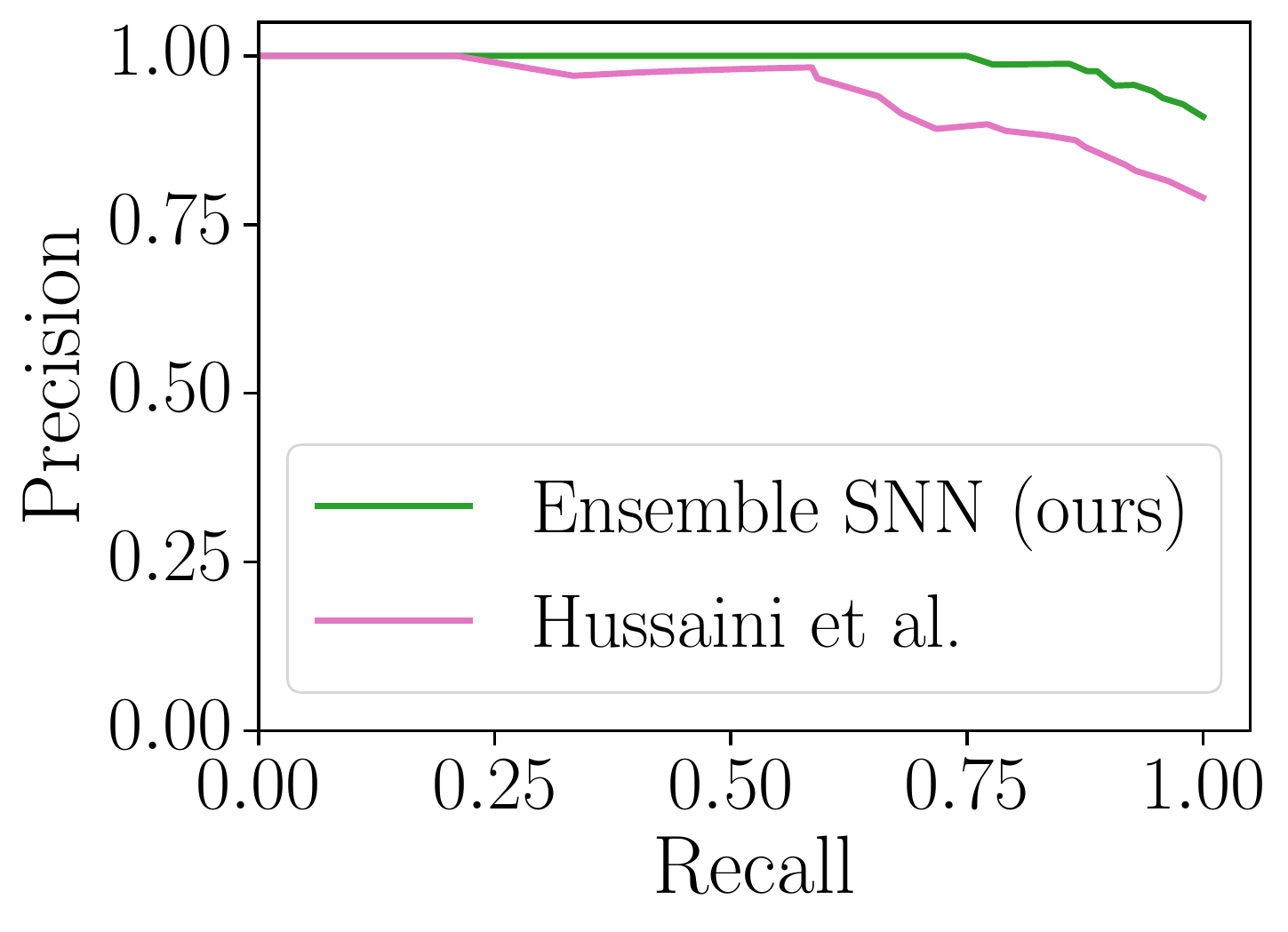}
    \vspace*{-0.2cm}
    \caption{Precision and recall curves of our proposed ensemble SNN in comparison to~\cite{Hussaini2022} in a small-scale environment.}
    \label{fig:A2A}
    \vspace*{-0.2cm}
\end{figure}

The P@100R of our proposed ensemble SNN model at 91.0\% is considerably higher than that of prior work on non-ensemble SNNs~\cite{Hussaini2022} (79.0\%), which in conjunction with the results in \Cref{comparison} demonstrates that our ensemble method is both scalable and provides improved performance in both small and large scale environments. %
The PR curve for these experiments is shown in \Cref{fig:A2A}.

\section{Conclusions and Future Work}

In this paper, we demonstrated that an ensemble of spiking neural networks can perform the visual place recognition task massively parallelized. Each ensemble member specializes in recognizing a small subset of places within a local region. Typically local ensemble members operate independently without any global regularization. We introduce global regularization by detecting and ignoring hyperactive neurons, which respond strongly to previously unseen places. Our experiments demonstrated significant performance gains and scalability improvements compared to prior SNNs, and comparable performance to NetVLAD, DenseVLAD and SAD. 

Future work will follow a number of research directions. We will investigate how our approach can be made robust to significant viewpoint changes. We are investigating to use event streams (from event cameras) as input data, instead of converting images to spike trains via rate coding, to further reduce the power requirements and localization latency. %
We are working towards implementing our method on Intel's neuromorphic processor, Intel Loihi \cite{davies2021advancing}. Deployment on neuromorphic hardware in similar applications \cite{frady2020neuromorphic} has demonstrated high energy efficiency, high throughput and low latency. 
Finally, we will investigate integrating our VPR system into a full SNN-based SLAM pipeline.

\IEEEtriggeratref{55}

\bibliographystyle{IEEEtran}
\bibliography{references}

\begin{thebibliography}{10}
\providecommand{\url}[1]{#1}
\csname url@rmstyle\endcsname
\providecommand{\newblock}{\relax}
\providecommand{\bibinfo}[2]{#2}
\providecommand\BIBentrySTDinterwordspacing{\spaceskip=0pt\relax}
\providecommand\BIBentryALTinterwordstretchfactor{4}
\providecommand\BIBentryALTinterwordspacing{\spaceskip=\fontdimen2\font plus
\BIBentryALTinterwordstretchfactor\fontdimen3\font minus
  \fontdimen4\font\relax}
\providecommand\BIBforeignlanguage[2]{{%
\expandafter\ifx\csname l@#1\endcsname\relax
\typeout{** WARNING: IEEEtran.bst: No hyphenation pattern has been}%
\typeout{** loaded for the language `#1'. Using the pattern for}%
\typeout{** the default language instead.}%
\else
\language=\csname l@#1\endcsname
\fi
#2}}

\bibitem{ghosh2009spiking}
S.~Ghosh-Dastidar and H.~Adeli, ``Spiking neural networks,'' \emph{Int. J.
  Neural Syst.}, vol.~19, no.~04, pp. 295--308, 2009.

\bibitem{gerstner2014neuronal}
W.~Gerstner, W.~M. Kistler, R.~Naud, and L.~Paninski, \emph{Neuronal dynamics:
  From single neurons to networks and models of cognition}.\hskip 1em plus
  0.5em minus 0.4em\relax Cambridge University Press, 2014.

\bibitem{frady2020neuromorphic}
E.~P. Frady \emph{et~al.}, ``Neuromorphic nearest neighbor search using
  {Intel's Pohoiki Springs},'' in \emph{Proc. Neuro-inspired Comput. Elements
  Worksh.}, 2020.

\bibitem{davies2021advancing}
M.~Davies \emph{et~al.}, ``Advancing neuromorphic computing with {Loihi}: A
  survey of results and outlook,'' \emph{Proc. IEEE}, vol. 109, no.~5, pp.
  911--934, 2021.

\bibitem{abadia2021cerebellar}
I.~Abad{\'\i}a, F.~Naveros, E.~Ros, R.~R. Carrillo, and N.~R. Luque, ``A
  cerebellar-based solution to the nondeterministic time delay problem in
  robotic control,'' \emph{Science Robotics}, vol.~6, no.~58, p. eabf2756,
  2021.

\bibitem{vitale2021event}
A.~Vitale, A.~Renner, C.~Nauer, D.~Scaramuzza, and Y.~Sandamirskaya,
  ``Event-driven vision and control for {UAVs} on a neuromorphic chip,'' in
  \emph{IEEE Int. Conf. Robot. Autom.}, 2021, pp. 103--109.

\bibitem{dupeyroux2021neuromorphic}
J.~Dupeyroux, J.~J. Hagenaars, F.~Paredes-Vall{\'e}s, and G.~C. de~Croon,
  ``Neuromorphic control for optic-flow-based landing of mavs using the loihi
  processor,'' in \emph{IEEE Int. Conf. Robot. Autom.}, 2021, pp. 96--102.

\bibitem{stagsted2020event}
R.~K. Stagsted, A.~Vitale, A.~Renner, L.~B. Larsen, A.~L. Christensen, and
  Y.~Sandamirskaya, ``Event-based pid controller fully realized in neuromorphic
  hardware: a one dof study,'' in \emph{IEEE/RSJ Int. Conf. Intell. Robot.
  Syst.}, 2020, pp. 10\,939--10\,944.

\bibitem{tieck2017towards}
J.~C.~V. Tieck \emph{et~al.}, ``Towards grasping with spiking neural networks
  for anthropomorphic robot hands,'' in \emph{Int. Conf. Artificial Neural
  Netw.}, 2017, pp. 43--51.

\bibitem{tieck2018controlling}
J.~C.~V. Tieck, L.~Steffen, J.~Kaiser, A.~Roennau, and R.~Dillmann,
  ``Controlling a robot arm for target reaching without planning using spiking
  neurons,'' in \emph{IEEE Int. Conf. Cogn. Informatics Cogn. Comput.}, 2018,
  pp. 111--116.

\bibitem{kreiser2020chip}
R.~Kreiser \emph{et~al.}, ``An on-chip spiking neural network for estimation of
  the head pose of the {iCub} robot,'' \emph{Front. Neurosci.}, vol.~14, no.
  551, 2020.

\bibitem{lele2021end}
A.~Lele, Y.~Fang, J.~Ting, and A.~Raychowdhury, ``An end-to-end spiking neural
  network platform for edge robotics: From event-cameras to central pattern
  generation,'' \emph{IEEE Trans. Cogn. Devel. Syst.}, 2021.

\bibitem{zhu2020spatio}
L.~Zhu, M.~Mangan, and B.~Webb, ``Spatio-temporal memory for navigation in a
  mushroom body model,'' in \emph{Conf. Biomimetic Biohybrid Syst.}, 2020, pp.
  415--426.

\bibitem{Hussaini2022}
S.~Hussaini, M.~Milford, and T.~Fischer, ``Spiking neural networks for visual
  place recognition via weighted neuronal assignments,'' \emph{IEEE Robot.
  Autom. Lett.}, vol.~7, no.~2, pp. 4094--4101, 2022.

\bibitem{Garg2021}
S.~Garg, T.~Fischer, and M.~Milford, ``{Where Is Your Place, Visual Place
  Recognition?}'' in \emph{Int. Joint Conf. Artif. Intell.}, 2021, pp.
  4416--4425.

\bibitem{Lowry2015}
S.~Lowry, N.~S{\"{u}}nderhauf, P.~Newman, J.~J. Leonard, D.~Cox, P.~Corke, and
  M.~J. Milford, ``{Visual place recognition: A survey},'' \emph{IEEE Trans.
  Robot.}, vol.~32, no.~1, pp. 1--19, 2015.

\bibitem{masone2021survey}
C.~Masone and B.~Caputo, ``A survey on deep visual place recognition,''
  \emph{IEEE Access}, vol.~9, pp. 19\,516--19\,547, 2021.

\bibitem{zhang2021visual}
X.~Zhang, L.~Wang, and Y.~Su, ``Visual place recognition: A survey from deep
  learning perspective,'' \emph{Pattern Recognition}, vol. 113, p. 107760,
  2021.

\bibitem{mountcastle1978organizing}
V.~Mountcastle, ``An organizing principle for cerebral function: the unit
  module and the distributed system,'' \emph{The Mindful Brain}, 1978.

\bibitem{krubitzer1995organization}
L.~Krubitzer, ``The organization of neocortex in mammals: are species
  differences really so different?'' \emph{Trends in Neurosciences}, vol.~18,
  no.~9, pp. 408--417, 1995.

\bibitem{varela2001brainweb}
F.~Varela, J.-P. Lachaux, E.~Rodriguez, and J.~Martinerie, ``The brainweb:
  phase synchronization and large-scale integration,'' \emph{Nature Reviews
  Neuroscience}, vol.~2, no.~4, pp. 229--239, 2001.

\bibitem{o2006modeling}
R.~C. O’Reilly, ``Modeling integration and dissociation in brain and
  cognitive development,'' \emph{Processes of Change in Brain and Cognitive
  Development: Attention and Performance}, vol.~21, pp. 375--402, 2006.

\bibitem{bock2014anatomical}
A.~S. Bock and I.~Fine, ``Anatomical and functional plasticity in early blind
  individuals and the mixture of experts architecture,'' \emph{Front. Human
  Neurosci.}, vol.~8, p. 971, 2014.

\bibitem{li2008aversive}
W.~Li, J.~D. Howard, T.~B. Parrish, and J.~A. Gottfried, ``Aversive learning
  enhances perceptual and cortical discrimination of indiscriminable odor
  cues,'' \emph{Science}, vol. 319, no. 5871, pp. 1842--1845, 2008.

\bibitem{jacobs1991adaptive}
R.~A. Jacobs, M.~I. Jordan, S.~J. Nowlan, and G.~E. Hinton, ``Adaptive mixtures
  of local experts,'' \emph{Neural Computation}, vol.~3, no.~1, pp. 79--87,
  1991.

\bibitem{happel1994design}
B.~L. Happel and J.~M. Murre, ``Design and evolution of modular neural network
  architectures,'' \emph{Neural Networks}, vol.~7, no. 6-7, pp. 985--1004,
  1994.

\bibitem{auda1999modular}
G.~Auda and M.~Kamel, ``Modular neural networks: a survey,''
  \emph{International journal of neural systems}, vol.~9, no.~02, pp. 129--151,
  1999.

\bibitem{Arandjelovic2018}
R.~Arandjelovic, P.~Gronat, A.~Torii, T.~Pajdla, and J.~Sivic, ``{NetVLAD: CNN}
  architecture for weakly supervised place recognition,'' \emph{IEEE Trans.
  Pattern Anal. Mach. Intell.}, vol.~40, no.~6, pp. 1437--1451, 2018.

\bibitem{torii201524}
A.~Torii, R.~Arandjelovic, J.~Sivic, M.~Okutomi, and T.~Pajdla, ``24/7 place
  recognition by view synthesis,'' in \emph{IEEE Conf. Comput. Vis. Pattern
  Recog.}, 2015, pp. 1808--1817.

\bibitem{milford2012seqslam}
M.~J. Milford and G.~F. Wyeth, ``{SeqSLAM}: Visual route-based navigation for
  sunny summer days and stormy winter nights,'' in \emph{IEEE Int. Conf. Robot.
  Autom.}, 2012, pp. 1643--1649.

\bibitem{sunderhauf2013we}
N.~S{\"u}nderhauf, P.~Neubert, and P.~Protzel, ``Are we there yet? {Challenging
  SeqSLAM} on a 3000 km journey across all four seasons,'' in \emph{IEEE Int.
  Conf. Robot. Autom. Worksh.}, 2013.

\bibitem{RobotCar}
W.~Maddern, G.~Pascoe, C.~Linegar, and P.~Newman, ``1 year, 1000 km: The
  {Oxford RobotCar} dataset,'' \emph{Int. J. Robot. Res.}, vol.~36, no.~1, pp.
  3--15, 2017.

\bibitem{sandamirskaya2022neuromorphic}
Y.~Sandamirskaya, M.~Kaboli, J.~Conradt, and T.~Celikel, ``Neuromorphic
  computing hardware and neural architectures for robotics,'' \emph{Science
  Robotics}, vol.~7, no.~67, p. eabl8419, 2022.

\bibitem{feldman2012spike}
D.~E. Feldman, ``The spike-timing dependence of plasticity,'' \emph{Neuron},
  vol.~75, no.~4, pp. 556--571, 2012.

\bibitem{ding2021optimal}
J.~Ding, Z.~Yu, Y.~Tian, and T.~Huang, ``Optimal {ANN-SNN} conversion for fast
  and accurate inference in deep spiking neural networks,'' in \emph{Int. Joint
  Conf. Artif. Intell.}, 2021.

\bibitem{rueckauer2017conversion}
B.~Rueckauer, I.-A. Lungu, Y.~Hu, M.~Pfeiffer, and S.-C. Liu, ``Conversion of
  continuous-valued deep networks to efficient event-driven networks for image
  classification,'' \emph{Front. Neurosci.}, vol.~11, p. 682, 2017.

\bibitem{bu2021optimal}
T.~Bu, W.~Fang, J.~Ding, P.~Dai, Z.~Yu, and T.~Huang, ``Optimal ann-snn
  conversion for high-accuracy and ultra-low-latency spiking neural networks,''
  in \emph{Int. Conf. Learn. Represent.}, 2021.

\bibitem{renner2021backpropagation}
A.~Renner, F.~Sheldon, A.~Zlotnik, L.~Tao, and A.~Sornborger, ``The
  backpropagation algorithm implemented on spiking neuromorphic hardware,''
  \emph{arXiv:2106.07030}, 2021.

\bibitem{lee2020enabling}
C.~Lee, S.~S. Sarwar, P.~Panda, G.~Srinivasan, and K.~Roy, ``Enabling
  spike-based backpropagation for training deep neural network architectures,''
  \emph{Front. Neurosci.}, vol.~14, no. 119, pp. 1--22, 2020.

\bibitem{tang2019spiking}
G.~Tang, A.~Shah, and K.~P. Michmizos, ``Spiking neural network on neuromorphic
  hardware for energy-efficient unidimensional {SLAM},'' in \emph{IEEE/RSJ Int.
  Conf. Intell. Robot. Syst.}, 2019, pp. 4176--4181.

\bibitem{tang2018gridbot}
G.~Tang and K.~P. Michmizos, ``Gridbot: an autonomous robot controlled by a
  spiking neural network mimicking the brain's navigational system,'' in
  \emph{Int. Conf. Neuromorphic Syst.}, 2018.

\bibitem{kreiser2018pose}
R.~Kreiser, A.~Renner, Y.~Sandamirskaya, and P.~Pienroj, ``Pose estimation and
  map formation with spiking neural networks: towards neuromorphic {SLAM},'' in
  \emph{IEEE/RSJ Int. Conf. Intell. Robot. Syst.}, 2018, pp. 2159--2166.

\bibitem{galluppi2012live}
F.~Galluppi \emph{et~al.}, ``Live demo: Spiking {RatSLAM: Rat} hippocampus
  cells in spiking neural hardware,'' in \emph{IEEE Conf. Biomed. Circuits
  Syst.}, 2012, p.~91.

\bibitem{milford2004ratslam}
M.~J. Milford, G.~F. Wyeth, and D.~Prasser, ``{RatSLAM}: a hippocampal model
  for simultaneous localization and mapping,'' in \emph{IEEE Int. Conf. Robot.
  Autom.}, 2004, pp. 403--408.

\bibitem{shim2016unsupervised}
Y.~Shim, A.~Philippides, K.~Staras, and P.~Husbands, ``Unsupervised learning in
  an ensemble of spiking neural networks mediated by {ITDP},'' \emph{PLoS
  Computational Biology}, vol.~12, no.~10, p. e1005137, 2016.

\bibitem{panda2017ensemblesnn}
P.~Panda, G.~Srinivasan, and K.~Roy, ``{EnsembleSNN: Distributed} assistive
  {STDP} learning for energy-efficient recognition in spiking neural
  networks,'' in \emph{Int. Joint Conf. Neural Networks}, 2017, pp. 2629--2635.

\bibitem{jegou2010vlad}
H.~J{\'e}gou, M.~Douze, C.~Schmid, and P.~P{\'e}rez, ``Aggregating local
  descriptors into a compact image representation,'' in \emph{IEEE Conf.
  Comput. Vis. Pattern Recog.}, 2010, pp. 3304--3311.

\bibitem{yu2019spatial}
J.~Yu, C.~Zhu, J.~Zhang, Q.~Huang, and D.~Tao, ``Spatial pyramid-enhanced
  netvlad with weighted triplet loss for place recognition,'' \emph{IEEE Trans.
  Neural Netw. Learn. Syst.}, vol.~31, no.~2, pp. 661--674, 2019.

\bibitem{hausler2021patch}
S.~Hausler, S.~Garg, M.~Xu, M.~Milford, and T.~Fischer, ``{Patch-NetVLAD}:
  Multi-scale fusion of locally-global descriptors for place recognition,'' in
  \emph{IEEE Conf. Comput. Vis. Pattern Recog.}, 2021, pp. 14\,141--14\,152.

\bibitem{khaliq2022multires}
A.~Khaliq, M.~Milford, and S.~Garg, ``{MultiRes-NetVLAD}: Augmenting place
  recognition training with low-resolution imagery,'' \emph{IEEE Robot. Autom.
  Lett.}, vol.~7, no.~2, pp. 3882--3889, 2022.

\bibitem{xu2021esa}
Y.~Xu, J.~Huang, J.~Wang, Y.~Wang, H.~Qin, and K.~Nan, ``{ESA-VLAD}: A
  lightweight network based on second-order attention and {NetVLAD} for loop
  closure detection,'' \emph{IEEE Robot. Autom. Lett.}, vol.~6, no.~4, pp.
  6545--6552, 2021.

\bibitem{yu2019neuroslam}
F.~Yu, J.~Shang, Y.~Hu, and M.~Milford, ``{NeuroSLAM: A} brain-inspired slam
  system for 3d environments,'' \emph{Biological Cybernetics}, vol. 113, no.~5,
  pp. 515--545, 2019.

\bibitem{chancan2020hybrid}
M.~Chanc{\'a}n, L.~Hernandez-Nunez, A.~Narendra, A.~B. Barron, and M.~Milford,
  ``A hybrid compact neural architecture for visual place recognition,''
  \emph{IEEE Robot. Autom. Lett.}, vol.~5, no.~2, pp. 993--1000, 2020.

\bibitem{nguyen2013spatio}
V.~A. Nguyen, J.~A. Starzyk, and W.-B. Goh, ``A spatio-temporal long-term
  memory approach for visual place recognition in mobile robotic navigation,''
  \emph{Robotics and Autonomous Systems}, vol.~61, no.~12, pp. 1744--1758,
  2013.

\bibitem{neubert2019neurologically}
P.~Neubert, S.~Schubert, and P.~Protzel, ``A neurologically inspired sequence
  processing model for mobile robot place recognition,'' \emph{IEEE Robot.
  Autom. Lett.}, vol.~4, no.~4, pp. 3200--3207, 2019.

\bibitem{newman2005slam}
P.~Newman and K.~Ho, ``{SLAM-loop} closing with visually salient features,'' in
  \emph{IEEE Int. Conf. Robot. Autom.}, 2005, pp. 635--642.

\bibitem{diehl2015unsupervised}
P.~U. Diehl and M.~Cook, ``Unsupervised learning of digit recognition using
  spike-timing-dependent plasticity,'' \emph{Front. Comput. Neurosci.}, vol.~9,
  no.~99, pp. 1--9, 2015.

\bibitem{stimberg2019brian}
M.~Stimberg, R.~Brette, and D.~F. Goodman, ``Brian 2, an intuitive and
  efficient neural simulator,'' \emph{Elife}, vol.~8, p. e47314, 2019.

\bibitem{molloy2020intelligent}
T.~L. Molloy, T.~Fischer, M.~Milford, and G.~N. Nair, ``Intelligent reference
  curation for visual place recognition via {Bayesian} selective fusion,''
  \emph{IEEE Robot. Autom. Lett.}, vol.~6, no.~2, pp. 588--595, 2020.

\bibitem{hausler2019multi}
S.~Hausler, A.~Jacobson, and M.~Milford, ``Multi-process fusion: Visual place
  recognition using multiple image processing methods,'' \emph{IEEE Robot.
  Autom. Lett.}, vol.~4, no.~2, pp. 1924--1931, 2019.

\end{thebibliography}

\end{document}